# Similarity-Driven Cluster Merging Method for Unsupervised Fuzzy Clustering

**Xuejian Xiong** [*]
Singapore-MIT Alliance
National University of Singapore
3 Science Drive 2
Singapore, 117543

**Kap Luk Chan**
School of EEE
Nanyang Technological University
Nanyang Avenue
Singapore 639798

**Kian Lee Tan**
Singapore-MIT Alliance
National University of Singapore
3 Science Drive 2
Singapore, 117543

## Abstract

In this paper, a similarity-driven cluster merging method is proposed for unsupervised fuzzy clustering. The cluster merging method is used to resolve the problem of cluster validation. Starting with an overspecified number of clusters in the data, pairs of similar clusters are merged based on the proposed similarity-driven cluster merging criterion. The similarity between clusters is calculated by a fuzzy cluster similarity matrix, while an adaptive threshold is used for merging. In addition, a modified generalized objective function is used for prototype-based fuzzy clustering. The function includes the $p$-norm distance measure as well as principal components of the clusters. The number of the principal components is determined automatically from the data being clustered. The properties of this unsupervised fuzzy clustering algorithm are illustrated by several experiments.

## 1 Introduction

In prototype-based fuzzy clustering methods, for example, the well-known Fuzzy $C$-Means (FCM) algorithm[Bezdek 1999], each cluster is represented by a prototypical point, known as the prototype. Each data point belongs to a cluster with a degree of likelihood as indicated by its fuzzy membership in the interval $[0, 1]$. Distance between a data point and a prototype is usually used as an optimizing measure in the objective function. Optimization is often performed by minimization of such distances over all the data points and prototypes. There are two main advantages of this objective-function based fuzzy clustering.

One is that the data points can be moved from one cluster to another to minimize the objective function. Another is that the knowledge about the shape or size of the clusters can be incorporated by using an appropriate distance measure in the objective function. However, several problems are still open for obtaining good performance from a fuzzy clustering algorithm. These concern the number of clusters in the data, uneven distribution of data points, initialization of the clustering algorithm, large variations of cluster's sizes, the shape of clusters, etc.

Determining the optimal number of clusters is an important issue in cluster validation for clustering. Traditionally, the optimal number of clusters is determined by evaluating a certain global validity measure of the $c$-partition for a range of $c$ values, and then picking the value of $c$ that optimizes the validity measure in some sense[Hammah 2000, Zahid 1998, Xie 1991, Bezdek 1974]. However, it is difficult to devise a unique measure that takes into account the variability in cluster shape, density, and size. Moreover, these procedures are computationally expensive because they require solving the optimization problem repeatedly for different values of the number of clusters $c$ over a pre-specified range $[c_{min}, c_{max}]$. In addition, the validity measures may not always give the correct number of clusters $c$ [Krishnapuram 1994].

In order to overcome these problems, researchers proposed merge-split or progressive clustering schemes based on the values of validity function[Krishnapuram 1992, Bezdek 1999]. Note that cluster splitting is the inverse approach of cluster merging, in which the data is treated as one cluster at the beginning. One cluster is split into two sub-clusters based on some assessment criteria. When there are no more clusters that should be split, the algorithm stops. We are more interested in cluster merging because it often requires less computation than cluster splitting. This is because when splitting a cluster, the new clusters' parameters need to be calculated using the data in the new clusters, while merging clusters, the parameters of the merged cluster can be obtained from the parameters of the original clusters.

---
[*]email: smaxx@nus.edu.sg, xiongxuejian@hotmail.com

Cluster merging[Krishnapuram 1992] is proposed as a way to select the number of clusters. The data is clustered by starting with an overspecified value of $c$. After the data is partitioned into $c$ clusters, similar clusters are merged together based on a given assessment criterion until no more clusters can be merged. The procedure of cluster validation is independent of the clustering algorithm, and the number of clusters is reduced dynamically. Krishnapuram *et al.* presented a compatible cluster merging method for unsupervised clustering[Krishnapuram 1994, Hoppner 1999]. Kaymak *et al.*[Kaymak 2002] also used the cluster merging method to determine the number of clusters in an extended FCM algorithm. The fuzzy inclusion measure is used to assess the similarity between two fuzzy clusters. Although an adaptive threshold is used, it cannot work well when the expected number of clusters in the data is larger than ten[Kaymak 2002].

The cluster merging approach offers an automatic and computationally less expensive way for cluster validation, but so far, most of the cluster merging methods heavily depend on the clustering procedure. In other words, these methods belong to dynamic cluster validation[Bezdek 1999]. They cannot be applied to other clustering algorithms easily. In the process, the intermediate clustering results are also affected by cluster merging. However, the static cluster validation method leads to heavy computation due to repeated clustering. To our knowledge, there are few works on cluster merging which combines the advantages of dynamic and static cluster validation approaches.

Therefore, in this paper, a similarity-driven cluster merging method is proposed for unsupervised fuzzy clustering, and it has advantages of both dynamic and static cluster validation. The proposed cluster merging method is based on a new similarity-driven cluster merging criterion. As a result, starting with a large number of clusters, pairs of similar clusters are repeatedly merged, until the correct number of clusters are determined. The similarity between clusters is calculated by a proposed fuzzy cluster similarity matrix. The merge threshold can be determined automatically and adaptively. Therefore, the over-partitioning of the data can be merged to the optimal fuzzy partitioning in a few steps. In addition, a modified generalized objective function is used for fuzzy clustering. The function includes the $p$-norm distance measure and the principal components of clusters. The number of the principal components is determined automatically from the data being clustered.

The organization of this paper is as follows. Section 2 presents the similarity-driven cluster merging method for solving the fuzzy cluster validity problem in unsupervised fuzzy clustering. In section 3, the modified generalized objective function based on the fuzzy $c$-prototype form is described. Experimental results on several data sets are presented in section 5.3. Finally, conclusion is given in section 6.

## 2 Similarity-Driven Cluster Merging Method

### 2.1 Similarity-Driven Cluster Merging Criterion

Let us consider a collection of data $\mathbf{X} = \{\mathbf{x} \in \Re^n\}$, in which there are $c$ clusters $\{\mathbf{P}_1, \mathbf{P}_2, \cdots, \mathbf{P}_c\}$. $\{\mathbf{V}_i \in \Re^n, i = 1, 2, \cdots, c\}$ are the prototypes of the corresponding clusters. If $dp_i$ is the fuzzy dispersion of the cluster $\mathbf{P}_i$, and $dv_{ij}$ denotes the dissimilarity between two clusters $\mathbf{P}_i$ and $\mathbf{P}_j$, then a fuzzy cluster similarity matrix $\mathbf{FR} = \{FR_{ij}, (i,j) = 1, 2, \cdots, c\}$ is defined as,

$$FR_{ij} = \frac{dp_i + dp_j}{dv_{ij}}. \quad (1)$$

The fuzzy dispersion $dp_i$ can be seen as a measure of the radius of $\mathbf{P}_i$, i.e. $dp_i = \sqrt{\frac{1}{n_i} \sum_{\mathbf{x} \in \mathbf{P}_i} \mu_i^m \parallel \mathbf{x} - \mathbf{V}_i \parallel^2}$, where $n_i$ is the number of data points in $\mathbf{P}_i$, $\mu_i = \{\mu_{i1}, \cdots, \mu_{iN}\}$ denotes the $i$-th row in the membership matrix $\mathbf{U} = \{\mu_{ij}\}$, and $m \in [0, \infty)$ is a fuzziness parameter. $dv_{ij}$ describes the dissimilarity between $\mathbf{P}_i$ and $\mathbf{P}_j$, i.e. $dv_{ij} = \parallel \mathbf{V}_i - \mathbf{V}_j \parallel$.

It can be seen that $FR_{ij}$ reflects the ratio of the sum of the fuzzy dispersion between two clusters, $\mathbf{P}_i$ and $\mathbf{P}_j$, to the distance between these two clusters. It can be concluded that $FR_{ij}$ satisfies the following conditions:

1. $FR_{ij} \geq 0$,
2. $FR_{ij} = FR_{ji}$,
3. If $dp_i = 0$ and $dp_j = 0$, then $FR_{ij} = 0$,
4. If $dp_j > dp_k$, and $dv_{ij} = dv_{ik}$, then $FR_{ij} > FR_{ik}$,
5. If $dp_j = dp_k$, and $dv_{ij} < dv_{ik}$, then $FR_{ij} > FR_{ik}$.

These conditions state that $FR_{ij}$ is nonnegative and symmetric. $FR_{ij}$ reflects the similarity between $\mathbf{P}_i$ and $\mathbf{P}_j$. Hence, it can be used to determine whether two clusters are similar or not, according to the following defined similarity-driven cluster merging criterion.

Considering a data set $\mathbf{X}$, there are $c$ clusters $\{\mathbf{P}_i, i = 1, 2, \cdots, c\}$. In each cluster, e.g. $\mathbf{P}_i$, $\mu_i$ is the membership vector of all data in $\mathbf{X}$ with respect to $\mathbf{P}_i$, and $\mathbf{V}_i$ denotes the prototype of $\mathbf{P}_i$. For a fuzzy similarity matrix $\mathbf{FR}$ and a given threshold $\tau$, the similarity-driven cluster merging criterion is defined as,

If $FR_{ij} \leq \tau$, two clusters $\mathbf{P}_i$ and $\mathbf{P}_j$ are completely separated;

If $FR_{ij} > \tau$, two clusters $\mathbf{P}_i$ and $\mathbf{P}_j$ are merged to form a new cluster $\mathbf{P}_{i'}$ with $\mu_{i'} = \mu_i + \mu_j$ and $\mathbf{V}_{i'} = \frac{\mathbf{V}_i + \mathbf{V}_j}{2}$,
then $c' = c - 1$. (2)

where $\mathbf{P}_{i'}$ refers to the new cluster after merging. $\mu_{i'}$ and $\mathbf{V}_{i'}$ denote the membership vector and the prototype of $\mathbf{P}_{i'}$, respectively. $c'$ is the number of clusters after merging. Note that the merging order of pairs of clusters in an iteration is according to the value of $FR_{ij}$ (see Table 1). Furthermore, a corresponding index is defined as $DB_{FR} = \frac{1}{c} \sum_{i=1}^{c} FR_i$, where $FR_i = \max_{i \neq j}\{FR_{ij}, (i, j = 1, 2, \cdots, c)\}$. The minimum $DB_{FR}$ corresponds to the optimal $c_{opt}$. Because $DB_{FR}$ is similar to the well-known DB index[Theodoridis 1999], it is named as the fuzzy DB index.

Table 1: The Merging Order Of Clusters In An Iteration, Based On The Similarity-Driven Cluster Merging Criterion.

| |
| --- |
| If $[i_1, j_1] = \arg\max_{(i,j)}\{FR_{ij} > \tau\}$, then clusters $\mathbf{P}_{i_1}$ and $\mathbf{P}_{j_1}$ are merged first; if $[i_2, j_2] = \arg\max_{(i \neq i_1, j \neq j_1)}\{FR_{ij} > \tau\}$, then clusters $\mathbf{P}_{i_2}$ and $\mathbf{P}_{j_2}$ are merged next; ...... if there is no $FR_{(i \neq \{i_1, i_2, \cdots\}, j \neq \{j_1, j_2, \cdots\})} > \tau$, then stop. |

### 2.2 Determination of Threshold for Similarity-Driven Cluster Merging Criterion

In order to define $\tau$, the following definition is given. For a data set $\mathbf{X} = \{\mathbf{x}_k, k = 1, \cdots, N\}$, $\mathbf{P} = \{\mathbf{P}_i, i = 1, \cdots, c\}$ is a set of $c$ clusters of $\mathbf{X}$, and the corresponding prototypes are $\{\mathbf{V}_i, i = 1, \cdots, c\}$. $\forall \mathbf{x}_k \in \mathbf{P}_i$, there is

$$\mathbf{P}'_i = \{\mathbf{x}_k | D(\mathbf{x}_k, \mathbf{V}_i) \leq dp_i, \quad \mathbf{x}_k \in \mathbf{P}_i, \quad k = 1, 2, \cdots, N\} \quad (3)$$

where $D(\mathbf{x}_k, \mathbf{V}_i)$ denotes the distance between $\mathbf{x}_k$ and $\mathbf{V}_i$, and $dp_i$ represents the fuzzy dispersion of $\mathbf{P}_i$.

It can be seen that $\mathbf{P}'_i \subset \mathbf{P}_i$. Nonetheless, $\mathbf{P}'_i$ can be used to represent the cluster $\mathbf{P}_i$, i.e. $\mathbf{P}'_i \approx \mathbf{P}_i$. Therefore, the following criteria can be obtained.

if $\mathbf{P}'_i \cap \mathbf{P}'_j = \emptyset$, i.e. $\sharp(\mathbf{P}'_i \cap \mathbf{P}'_j) = 0$,
then $dp_i + dp_j < dv_{ij}$ i.e. $FR_{ij} < 1$; (4)
if $\mathbf{P}'_i \cap \mathbf{P}'_j \neq \emptyset$, i.e. $\sharp(\mathbf{P}'_i \cap \mathbf{P}'_j) \geq 1$
then $dp_i + dp_j \geq dv_{ij}$ i.e. $FR_{ij} \geq 1$. (5)

where $\sharp(\mathbf{P}_i)$ denotes the number of data points in the cluster $\mathbf{P}_i$.

The contour of the dispersion of a cluster can be drawn to represent the cluster as shown in Figure 1. If two clusters, $\mathbf{P}_i$ and $\mathbf{P}_j$, are far away from each other, i.e. there is no intersection between two dispersion contours (refer to equation(4)), it is believed that the two clusters are well separated from each other. As shown in Figure 1, $\mathbf{P}_1$ and $\mathbf{P}_4$ are two completely separated clusters. If there is an intersection between the dispersion contours of two clusters, it can be said that these two clusters are overlapped clusters and should be merged together (refer to equation(5)). From Figure 1, it can be considered that $\mathbf{P}_2$ and $\mathbf{P}_3$, $\mathbf{P}_4$ and $\mathbf{P}_5$, $\mathbf{P}_5$ and $\mathbf{P}_6$ are overlapped with each other. However, if two dispersion contours are at tangent, i.e. $dp_i + dp_j = dv_{ij}$ and then $FR_{ij} = 1$, it can be considered that $\mathbf{P}_i$ and $\mathbf{P}_j$ are separated. Therefore, the similarity threshold $\tau$ can be fixed as 1. $\tau$ can also be given other values. If $\tau > 1$, for example $\tau = 2$, it means that two clusters can be seen as separating from each other well even though they overlap much more. Otherwise, if $\tau < 1$, for example $\tau = 0.5$, it means that two clusters should be merged together even though they are well separated.

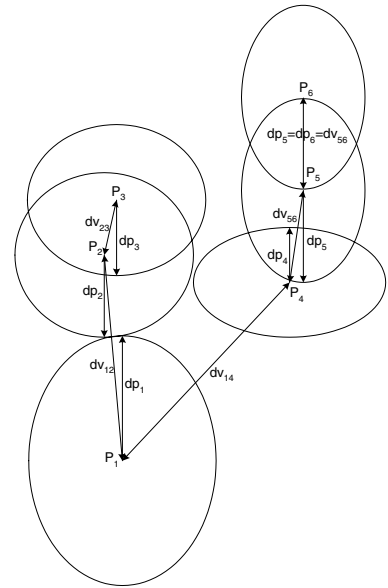

Figure 1: Intersection Between Pairs Of Clusters Represented By Their Dispersion Contours.

The value of $\tau$ can affect the final solution and speed of the cluster merging. Thus, the definition of the similarity-driven cluster merging criterion, in equation (2), can be refined as follows,

$$
\begin{aligned}
&\text{If } FR_{ij} < \tau_1, && \text{two clusters } \mathbf{P}_i \text{ and } \mathbf{P}_j \text{ are} \\
& && \text{completely separated;} \\
&\text{If } \tau_1 \leq FR_{ij} \leq \tau_2, && \text{an annealing technique is} \\
& && \text{applied to find the appropriate} \\
& && \tau \text{ for equation(2);} \\
&\text{If } FR_{ij} > \tau_2, && \text{two clusters } \mathbf{P}_i \text{ and } \mathbf{P}_j \text{ are} \\
& && \text{merged to form a new cluster} \\
& && \mathbf{P}_{i'} \text{ with } \mu_{i'} = \mu_i + \mu_j \text{ and} \\
& && \mathbf{V}_{i'} = \tfrac{\mathbf{V}_i + \mathbf{V}_j}{2}, \\
& && \text{then } c' = c - 1. \quad (6)
\end{aligned}
$$

Based on the discussion of equations (4) and (5) and Figure 1, it is seen that $\tau_1$ can be reasonably set as 1. Normally, if $FR_{ij} \geq 2$, $\mathbf{P}_i$ and $\mathbf{P}_j$ will be considered as the overlapped clusters to be merged with no doubt. As a result, $\tau_2$ is set to 2. If $1 \leq FR_{ij} \leq 2$, the appropriate value of the threshold is obtained adaptively and automatically by using an annealing technique.

## 3   A Modified Generalized Objective Function

A modified generalized objective function for the unsupervised fuzzy clustering algorithm is described in this section. The function consists of the *p*-norm distance measure and principal components of clusters.

Consider a collection of $N$ data $\{\mathbf{x}_k \in \Re^n, k = 1, 2, \cdots, N\}$ forming the data set $\mathbf{X}$. There are $c$ clusters whose prototypes are $\mathbf{V} = \{\mathbf{V}_i \in \Re^n, i = 1, \cdots, c\}$. The modified generalized objective function based on [Bezdek 1999, Yoshinari 1993] is proposed as follows,

$$J_{\{m,p\}}(\mathbf{U},\mathbf{V};\mathbf{X}) \triangleq \sum_{i=1}^{c}\sum_{k=1}^{N}(\mu_{ik})^m\{D_p(ik)+gD_r(ik)\}$$
$$=\sum_{i=1}^{c}\sum_{k=1}^{N}(\mu_{ik})^m\left\{\{\|\mathbf{x}_k-\mathbf{V}_i\|_p\}^p+g\sum_{s=1}^{r}\mathbf{S}_{is}^T(\mathbf{x}_k-\mathbf{V}_i)\right\} \quad (7)$$

where, $p \geq 1$, $m \in [0, \infty)$ is a fuzziness parameter, and $g \in [0,1]$ is a weight. $\{\mathbf{S}_{is} \in \Re^n, s = 1, \cdots, r\}$ are $r$ eigenvectors of the generalized within-cluster scatter matrix of the cluster $\mathbf{P}_i$. $\mathbf{U} = \{\mu_{ik}\}$ is the fuzzy membership matrix, and $\mu_{ik}$ should satisfy the following constraints:

$$
\begin{aligned}
0 \leq\quad & \mu_{ik} &&\leq 1 \quad \forall \quad i,k, \\
& \textstyle\sum_{i=1}^{c}\mu_{ik} &&= 1 \quad \forall \quad k, \\
0 <\quad & \textstyle\sum_{k=1}^{N}\mu_{ik} &&< N \quad \forall \quad i. \quad (8)
\end{aligned}
$$

The first term $D_p(ik)$, in the objective function $J_{\{m,p\}}(\mathbf{U},\mathbf{V};\mathbf{X})$, characterizes the distance from a data point $\mathbf{x}_k$ to the cluster $\mathbf{P}_i$, based on the *p*-norm distance measure. The second term $D_r(ik)$ introduces the principal axes of the cluster $\mathbf{P}_i$, which are determined by the collection of $r > 0$ linearly independent vectors $\{\mathbf{S}_{is}, s = 1, 2, \cdots, r\}$. $\{\mathbf{S}_{i1}, \mathbf{S}_{i2}, \cdots, \mathbf{S}_{ir}\}$ are eigenvectors corresponding to the first $r$ largest eigenvalues of the generalized within-cluster scatter matrix $\mathbf{E}_i = \sum_{k=1}^{N}(\mu_{ik})^m(\mathbf{x}_k - \mathbf{V}_i)(\mathbf{x}_k - \mathbf{V}_i)^T$.

$\{\mathbf{S}_{is}, s = 1, 2, \cdots, r\}$ gives the cohesiveness of the cluster $\mathbf{P}_i$. In fact, $\{\mathbf{S}_{is}, s = 1, 2, \cdots, r\}$ are the $r$ principal eigenvectors of the cluster $\mathbf{P}_i$. They give the most important directions, along which most of the data points in the cluster scatter. Through the weighted term $D_r(ik)$, the principal directions of the cluster $\mathbf{P}_i$ can be emphasized. In other words, the search for the prototype $\mathbf{V}_i$ is only along the principal directions. As a result, the speed of the search is improved. Especially for a large number of data points, the appropriate value of $r$ can be selected to significantly improve the convergence speed of the fuzzy clustering algorithm.

Choosing a suitable value of $r$ in different applications is still a problem. For the fuzzy *c*-elliptotypes and fuzzy *c*-variants algorithms, two variations of the FCM[Bezdek 1999, Yoshinari 1993], $r$ must be specified *a priori* based on the assumed shape of clusters. However, it is difficult to imagine the shape of clusters if the dimension of the data is larger than three, i.e. $n > 3$. Since the minimum description length (MDL)[Hyvarinen 2001] is one of the well-known criteria for model order selection, the MDL is used here to find the optimal value of $r$. For $N$ input data $\{\mathbf{x}_k \in \Re^n, k = 1, 2, \cdots, N\}$, there is

$$MDL(j) = -(n-j)N \ln \frac{\mathcal{G}(\lambda_{j+1},\cdots,\lambda_n)}{\mathcal{A}(\lambda_{j+1},\cdots,\lambda_n)} + \frac{1}{2}j(2n-j)\ln N \quad (9)$$

where $\lambda_1 \geq \lambda_2 \geq \cdots \geq \lambda_n$ denote the eigenvalues of $\mathbf{E}_i$, and $j \in [1, 2, \cdots, n]$. $\mathcal{G}(\cdot)$ and $\mathcal{A}(\cdot)$ denote the geometric mean and the arithmetic mean of their arguments, respectively. Hence, the optimal value of $r$ can be determined as follows,

$$r = \{j | \min_{j=r_1, r_1+1, \cdots, n-1} MDL(j)\} \quad (10)$$

That is, equation (10) searches for the optimal $r$ from $[r_1, \cdots, n-1]$. Normally, $r_1 = 1$.

## 4 The Complete Unsupervised Fuzzy Clustering Algorithm

The unsupervised fuzzy clustering algorithm consists of a modified generalized objective function for fuzzy clustering, and a similarity-driven cluster merging criterion for cluster merging, i.e. the GFC-SD algorithm in short. The complete GFC-SD algorithm is described step by step as follows:

**step 1.** *Initialization*:
Pre-selecting the maximum value for the number of clusters $c = c_{max}$, obviously $c_{max} < N$; pre-defining $g$, $p$, $r_1$, $m$, the tolerance $\epsilon$, the merging thresholds $\tau_1$ and $\tau_2$; setting the initial membership matrix $\mathbf{U}$ subject to constraints in equation (8).

**step 2.** *Updating*:
Updating the cluster prototypes $\mathbf{V}$ and the membership matrix $\mathbf{U}$. The updating formulae can be obtained by differentiating the generalized objective function $J_{\{m,p\}}(\mathbf{U}, \mathbf{V}; \mathbf{X})$ with respect to $\mathbf{V}$ and $\mathbf{U}$, respectively.

**step 3.** *The penalty rule*:
If the given stopping criterion is satisfied, i.e. $\| \mathbf{U}_{new} - \mathbf{U} \| < \epsilon$, go to next step, else go back to step 2, replace the old $\mathbf{U}$ with the new partition matrix $\mathbf{U}_{new}$.

**step 4.** *Cluster merging*:
Merging clusters based on proposed similarity-driven cluster merging criterion. If $c$ is not changed, then stop the procedure, else go back to step 2, repeat the whole procedure according to the new number of clusters $c$, and use current corresponding $\mathbf{V}$ and $\mathbf{U}$ as the initialization.

## 5 Experiments

In this section, the performance of the GFC-SD algorithm is studied. For comparison, the GFC-SD algorithm is applied to an artificially generated two-dimensional data set, which was used in [Kaymak 2002]. Moreover, the well-known IRIS data set from the UCI Machine Learning Repository is classified based on the clustering results of the GFC-SD algorithm. Finally, a gene expression data set is studied by using the GFC-SD algorithm. All experiments are done with a 2-norm distance measure, i.e. $p = 2$. The tolerance for fuzzy clustering $\epsilon$ is selected as 0.001. The merging threshold $\tau$ is determined adaptively according to equation (6) with $\tau_1 = 1$ and $\tau_2 = 2$. All experimental results are obtained on a 1.72GHz Pentium IV machine with 256MB memory, running Matlab 5.3 on Windows XP.

### 5.1 The Artificial Data Set with Uneven-Distributed Groups

As mentioned in [Kaymak 2002], four groups of data are generated randomly from normal distributions around four centers given in Table 2. The number of sample points in each group is also indicated. It can be seen that the number of sample points in group 1 is much larger than that of other three groups. That is, the differences in cluster density are quite large.

Table 2: The Group Centers And Number Of Samples In Each Group Of The Artificial Data Set With Uneven-Distributed Groups.

| group | 1 | 2 | 3 | 4 |
|---|---|---|---|---|
| original center | (-0.5,-0.4) | (0.1,0.2) | (0.5,0.7) | (0.6,-0.3) |
| number of samples | 300 | 30 | 30 | 50 |

In this experiment, the goal is to automatically detect clusters reflecting the underlying structure of the data set. The well-known FCM method with the popular Xie's cluster validity function[Xie 1991], i.e. FCM-Xie in short, is used for comparison. By using the FCM-Xie, the number of clusters $c$ is determined based on the minimal value of the Xie's cluster validity function. Here, the range of values of $c$ is $[2, 20]$. From Figure 2(a), it can be observed that the conventional approach, FCM-Xie, detects $c = 2$. It fails to determine the correct number of clusters in the data due to the largely uneven distribution of the data. In addition, in [Kaymak 2002], Kaymak's extended FCM algorithm also cannot find the correct $c$ of this largely uneven-distributed data set. The proposed GFC-SD algorithm, however, detects the four groups present in the data correctly, as shown in Figure 2(b). Hence, the GFC-SD algorithm is more robust for largely uneven-distributed data than the FCM-Xie algorithm, as well as Kaymak's extended FCM algorithm.

Like the experimental procedure in [Kaymak 2002], the influence of initialization on the GFC-SD algorithm is also studied. The data set is clustered 1000 times with the FCM and the GFC-SD algorithms, respectively. At each time, the randomly initialized fuzzy partitions, $\mathbf{U}$, are input into the algorithms. The FCM algorithm is set to partition the data into four clusters, i.e. $c = 4$, while the GFC-SD algorithm is started with twenty clusters, i.e. $c_{max} = 20$. After 1000 experiments, the mean and standard deviation of obtained cluster prototypes are shown in Table 3. Obviously, the cluster prototypes found by the GFC-SD algorithm are closer to the true centers than those found by the FCM algorithm. Moreover, the standard deviation of the GFC-SD found prototypes is much more lower. In fact, it almost equals to zero. The FCM algorithm has difficulty with small data groups, whose prototypes will be attracted by those of large ones. If there are much more data points in the large group than those in the small group, the latter one will be missed when bad initialization is given. Therefore, its obtained mean cluster prototype is far away from the true center and the corresponding standard deviation is very large. It can be concluded that the GFC-SD algorithm is much more robust to the initialization.

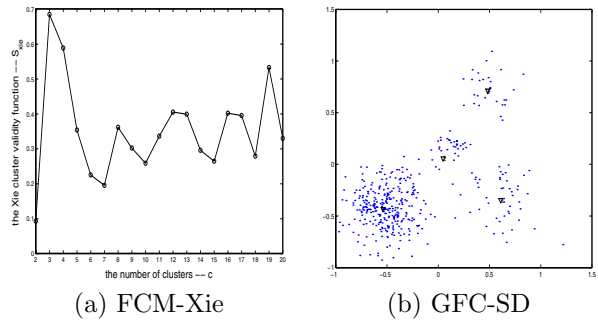

(a) FCM-Xie    (b) GFC-SD

Figure 2: (a) The FCM-Xie algorithm fails in determining the four clusters in the data set. (b) The GFC-SD algorithm automatically detects the correct number of clusters in the data set. The searched GFC-SD prototypes are denoted by the black triangle and numbers.

Table 3: Mean And Standard Deviation Of Cluster Prototypes Found By The FCM and GFC-SD Algorithms After 1000 Experiments With Random Initialization.

|   | FCM prototype | | GFC-SD prototype | |
|---|---|---|---|---|
|   | mean | std. dev. | mean | std. dev. |
| 1 | (-0.59,-0.42) | (0.019,0.081) | (-0.54,-0.43) | $< 10^{-13}$ |
| 2 | (-0.41,-0.39) | (0.007,0.117) | (0.05,0.06) | $< 10^{-13}$ |
| 3 | (0.42,0.61) | (0.007,0.010) | (0.48,0.71) | $< 10^{-13}$ |
| 4 | (0.58,-0.28) | (0.003,0.004) | (0.61,-0.35) | $< 10^{-13}$ |

To compare the computational load of various algorithms, different algorithms have been run 1000 times (listed in Table 4). Similarly, the algorithms are initialized randomly at each time. Here, GFC means the fuzzy clustering algorithm only with generalized objective function. For $c = 4$, the computational load of the GFC algorithm is larger than that of the FCM algorithm because of the additional calculation of the second term in the generalized objective function. However, by using the merging method to find the optimal partitions, i.e. GFC-SD, the computational load is only half of that using the conventional FCM-Xie approach (see Table 4).

## 5.2 The IRIS Data

The IRIS data, from the UCI Machine Learning Repository, contains three classes and 50 samples in each class, where each class refers to a type of iris plant: Iris Setosa, Iris Versicolour, or Iris Virginica. One class is linearly separable from the other two; the latter two are not linearly separable from each other. The dimension of each IRIS datum is four, i.e. $n = 4$.

By using the FCM-Xie algorithm, the optimal number of clusters is two and $c = 3$ is only sub-optimal (Figure 3). This result does not match the real structure of the IRIS data. Therefore, the correct clusters cannot be found automatically by using the conventional FCM-Xie algorithm.

For the GFC-SD algorithm, the clustering starts with $c_{max} = 20$, and the optimal number of clusters, $c = 3$, is obtained in six iterations. The overall accuracy of unsupervised classification based on the clustering results of GFC-SD is 93.33%. Table 5 provides the confusion matrix of this classification results.

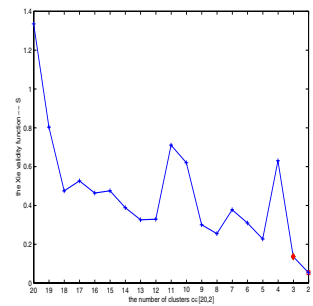

Figure 3: The FCM-Xie Algorithm Cannot Detect The Real Structure Of the IRIS Data.

Table 4: Average Computational Load Over 1000 Times For Various Clustering Algorithm.

|   | FCM | FCM-Xie | GFC | GFC-SD |
|---|---|---|---|---|
| c | 4 | [2,20] | 4 | 20 |
| time(s) | 7.96 | 467.41 | 11.17 | 243.09 |

Table 5: Unsupervised Classification Results Based On The GFC-SD Clustering Results Of The IRIS Data.

|   | Classified by GFC-SD | | | |
|---|---|---|---|---|
| original | class 1 | class 2 | class 3 | total |
| class 1 | 50 | 0 | 0 | 50 |
| class 2 | 0 | 48 | 2 | 50 |
| class 3 | 0 | 8 | 42 | 50 |
| total | 50 | 56 | 44 | 150 |

### 5.3 Gene Expression Data

The proposed GFC-SD algorithm is applied to study a gene expression data set, i.e. the serum data set. The serum data[Iyer 1999] contains expression levels of 8613 human genes by studying the response of human fibroblasts to serum. A subset of 517 genes whose expression levels changed substantially across samples was analyzed in [Dembele 2003, Sharan 2000, Eisen 1998, Iyer 1999]. Therefore, the serum data, consisting of 517 genes whose expression level is obtained from 13 experiments, are used here. All gene expression data are preprocessed in the same way as in [Dembele 2003] by using the variance normalization. This is done by subtracting its mean across the experiments from the expression level of each gene, e.g. $\overline{x}_k$, and dividing by the standard deviation across the experiments,

$$x'_{kj} = \frac{x_{kj} - \overline{x}_k}{\sqrt{\frac{1}{n}\sum_{j=1}^{n}(x_{kj} - \overline{x}_k)^2}} \quad (11)$$

where $n$ is the number of experiments.

To evaluate the performance of the proposed GFC-SD algorithm, the FCM-Xie algorithm used in [Dembele 2003, Dougherty 2002] is also applied here for clustering the gene expression data sets. In this experiment, the fuzziness parameter $m$ is selected as 1.25, which follows the empirical method proposed in [Dembele 2003].

Figure 4 presents the clustering results from using the proposed GFC-SD and the FCM-Xie algorithms. It can be observed from Figure 4(a) that, starting with 30 clusters, the number of clusters is reduced to 25, 22, 20, 17, 15, 13, 11, and finally 10 in only nine steps, based on the proposed similarity-driven cluster merging method. As a result, the number of clusters $c$ is determined as 10, which also corresponds to the minimal value of $DB_{FR}$. For using the FCM-Xie, the number of clusters can only be found after the exhausting search from all possible values of $c$. In this case, the range of $c$ is from 2 to 30. After 29 clustering iterations, in Figure 4(b), the number of clusters is fixed as two referring to the minimum $S_{xie}$. In [Dembele 2003, Sharan 2000, Iyer 1999], it is consistently agreed that there are 10 clusters in the serum data set with 517 genes. As a result, the proposed GFC-SD algorithm is effective for finding the number of gene clusters automatically and correctly.

Obviously, repeated clustering leads to a heavy computation, especially for gene expression data which have high dimensionality and a large number of genes. The consumed time for running the GFC-SD and FCM-Xie is $1.1911 \times 10^2$ seconds and $3.1029 \times 10^3$ seconds, respectively. It can be seen that running the FCM-Xie takes almost 30 times longer than running the GFC-SD. Furthermore, if the given $c_{max}$ is increased, e.g. $c_{max} = 40$, the time gap between these two algorithms will be enlarged significantly.

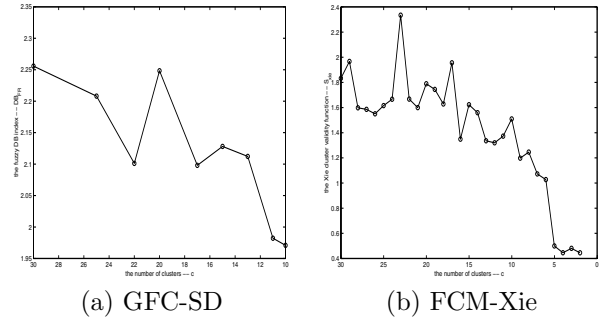

(a) GFC-SD      (b) FCM-Xie

Figure 4: The Number Of Clusters Of The Serum Data Is Determined As Ten And Two, By Using The Proposed GFC-SD And The FCM-Xie Algorithms, Respectively.

Table 6: The Number Of Clusters $c$ And The Number of Principal Components $r$ Of The Serum Clusters In Each Clustering Iteration.

| iteration | 1 | 2 | 3 | 4 | 5 | 6 | 7 | 8 | 9 |
|---|---|---|---|---|---|---|---|---|---|
| $c$ | 30 | 25 | 22 | 20 | 17 | 15 | 13 | 11 | 10 |
| $r$ | 8 | 9 | 10 | 10 | 11 | 11 | 12 | 11 | 11 |

An additional advantage of the proposed GFC-SD algorithm is that the optimal value of $r$ is found automatically (refer to equation (10)). Therefore, the number of principal components of each cluster can be adaptively determined. For the serum data, the values of $r$ and $c$ in each clustering iteration are listed in Table 6. It is observed that there are around ten principal components constructing the serum clusters. Therefore, the GFC-SD algorithm can perform feature selection of gene expression data to some extent.

## 6 Conclusion

In this paper, a similarity-driven cluster merging method is proposed for unsupervised fuzzy clustering. The cluster merging method is used to resolve the problem of cluster validation. The data is clustered initially with an overspecified number of clusters. Pairs of similar clusters are merged based on the proposed similarity-driven cluster merging criterion. The similarity between clusters is calculated by a fuzzy cluster similarity matrix, while an adaptive threshold is used for merging. Therefore, only a few iterations are needed to find the optimal number of clusters $c$, and more precise partitions can be obtained. Moveover, the dependency of the clustering results on the random initialization is reduced. For prototype-based fuzzy clustering, a modified generalized objective function is used. The function introduces the principal components of clusters by including an additional term. Because the data are grouped into different clusters along the principal directions of the clusters, the computational precision can be improved while the computation time can be reduced.

Several data sets are used to evaluate the performance of the GFC-SD algorithm. It can be concluded from the experiments that clustering using the GFC-SD algorithm is far less sensitive to initialization and more reliable than the compared methods. Moreover, because the partitions after one merging step are always the initialization of the next iteration of clustering, the total time of the fuzzy clustering is reduced. Thus, by using the GFC-SD algorithm, the optimal number of clusters and the optimal partitions of the data set can be obtained in relatively fewer iterations.